\documentclass[preprint, 3p, twocolumn]{elsarticle}

\usepackage{times}
\usepackage{latexsym}
\usepackage{graphicx}
\usepackage{amssymb}
\usepackage{float}

\usepackage{lineno,hyperref}
\usepackage{color,soul}
\usepackage[utf8]{inputenc}
\usepackage{xcolor}
\usepackage{listings}
\usepackage{multirow}
\usepackage{colortbl}
\usepackage{setspace}
\usepackage{comment}
\usepackage{url}
\usepackage{hyperref}
\usepackage{booktabs} 
\usepackage{tabu} 
\usepackage{tabularx}
\usepackage[T1]{fontenc}
\usepackage[scaled=0.9]{beramono}
\usepackage{microtype}
\usepackage[justification=centering]{caption}

\usepackage{enumitem}
\setitemize{noitemsep,,parsep=0pt,partopsep=0pt}
\usepackage{rotating}
\usepackage{courier}
\usepackage{newfloat}
\DeclareCaptionType{InfoBox}
\usepackage{tabularx}

\modulolinenumbers[5]

\journal{Journal of \LaTeX\ Templates}

\begin{document}

\begin{frontmatter}

%\title{Understanding and Measuring the Impact of Graph Embeddings in Scientific Text}
%\title{On the Impact of Knowledge-based Linguistic Analysis in the Quality of Scientific Embeddings}
\title{On the Impact of Knowledge-based Linguistic Annotations in the Quality of Scientific Embeddings}
%\tnoteref{mytitlenote} (12-18 pages, Deadline May 30th)}
%\tnotetext[mytitlenote]{Obviously this is not the definitive title}


\author[expertsystemaddress]{Andres Garcia-Silva\corref{mycorrespondingauthor}} 
\ead{agarcia@expert.ai}

\author[expertsystemaddress]{Ronald Denaux\corref{mycorrespondingauthor}} 
\ead{rdenaux@expert.ai}

\author[expertsystemaddress]{Jose Manuel Gomez-Perez\corref{mycorrespondingauthor}}
\ead{jmgomez@expert.ai}

\cortext[mycorrespondingauthor]{Corresponding author} 
\address[expertsystemaddress]{Expert.ai Research Lab, Prof. Waskman 10, 28036 Madrid, Spain}

\begin{abstract}
In essence, embedding algorithms work by optimizing the distance between a word and its usual context in order to generate an embedding space that encodes the distributional representation of words. In addition to single words or word pieces, other features which result from the linguistic analysis of text, including lexical, grammatical and semantic information, can be used to improve the quality of embedding spaces. However, until now we did not have a precise understanding of the impact that such individual annotations and their possible combinations may have in the quality of the embeddings. In this paper, we conduct a comprehensive study on the use of explicit linguistic annotations to generate embeddings from a scientific corpus and quantify their impact in the resulting representations. Our results show how the effect of such annotations in the embeddings varies depending on the evaluation task. In general, we observe that learning embeddings using linguistic annotations contributes to achieve better evaluation results. 

\end{abstract}


\end{frontmatter}

%\linenumbers

\section{Introduction}
Distributed word representations in the form of dense vectors, known as word embeddings, have become extremely popular for NLP tasks such as part-of-speech tagging, chunking, named entity recognition, semantic role labeling and synonym detection. Such interest was significantly sparked by the word2vec algorithm proposed in~\cite{mikolov2013a}, which provided an efficient way to learn word embeddings from large corpora based on word context and negative sampling, and continues to date. %Originally proposed in~\cite{baroni2010distmem}, 
Ample research has been put into the development of increasingly effective means to produce word embeddings, resulting in algorithms like GloVe~\cite{pennington2014glove}, Swivel~\cite{shazeer2016swivel} or fastText~\cite{bojanowski2016enriching}. Lately, much of this effort has been focused on neural language models that produce contextual word representations at a scale never seen before, like ELMo~\cite{PetersELMO2018}, BERT~\cite{devlin2018bert} and XLNet~\cite{Yang2019}. 

In general, word embedding algorithms are trained to minimize the distance between a token and its usual context of neighboring tokens in a document corpus. The set of tokens that occur in a text after some tokenization is applied may include e.g. words ("making"), punctuation marks (";"), multi-word expressions ("day-care center") or combinations of words and punctuation marks ("However,"). Indeed, the information that is actually encoded in the resulting embeddings depends not only on the neighboring context of a word or its frequency in the corpus, but also on the strategies used to chunk the text into sequences of individual tokens. 

In Table~\ref{tab:exampleTokenizations}, we show several examples of tokenizations that can be produced by processing a sentence following different tokenization strategies. In the example, the original text is annotated with information on space-separated tokens (\textit{t}) and surface forms (\textit{sf}). Additional linguistic and semantic annotations can also be derived from the text, like the lemma (\textit{l}) associated to each token, the corresponding sense or concept\footnote{Note that in the example, concept information is linked to a knowledge graph through a unique identifier.} extracted through word-sense disambiguation (\textit{c}), and grammatical information about the role each token plays in the context of a particular sentence (\textit{g}), like part-of-speech.  

\begin{table*}[t]
\small
  \caption{\small Tokenizations for the first window of size W=3 for the sentence: ``With regard to breathing in the locus coeruleus,''. First we show basic space-based tokenization. Next, sequences for surface forms, lemmas, syncons and grammar type.}
  \label{tab:exampleTokenizations}
\begin{center}  
\begin{tabular}{lccccccccc}
\hline
context & \(t^{i-3}\) & \(t^{i-2}\) & \(t^{i-1}\) & \(t^{i}\) & \(t^{i+1}\) & \(t^{i+2}\) & \(t^{i+3}\) & \\
\hline
\textit{t} & With & regard & to & breathing & in & the &  locus & coeruleus,\\
\textit{sf} & With regard to & breathing  & in & the & locus coeruleus & $\varnothing$\\
\textit{l} & with regard to & breathe  & $\varnothing$ & $\varnothing$ & locus coeruleus & $\varnothing$\\
\textit{c} & en\#216081 & en\#76230 & $\varnothing$ & $\varnothing$ & en\#101470452 & $\varnothing$ \\
\textit{g} & PRE & VER & PRE & ART & NOU & PNT  \\
\hline
\end{tabular}
\end{center}
\end{table*}

The resulting sequences of tokens used to train the embeddings differ significantly both in number of tokens and complexity. This can have a decisive impact on what information is actually encoded and consequently on the quality of the embeddings. In fact, the analysis of optimal tokenization strategies to train word embeddings working at either word (Word2Vec), character (ELMo), byte-pair encoding (BPEmb~\cite{heinzerling-strube-2018-bpemb}) or word-piece (BERT) levels is already well covered in the literature. However, none of these approaches investigate the role that explicit linguistic and semantic annotations (henceforth, linguistic annotations) like the ones above-mentioned, as well as their combinations, can play in the generation of the embeddings. 

In this paper, we apply this observation to investigate how linguistic annotations extracted from the text can be used to produce enriched, higher-quality embeddings. In doing so, we focus on the scientific domain as a field with a significantly complex terminology, rich in multi-word expressions and other linguistic artifacts. Scientific resources, such as Springer Nature's SciGraph~\cite{HammondPT17:SciGraphModeling} which contains over 3M scholarly publications, can be considerably rich in domain-specific multi-word expressions. This is particularly noticeable in Table~\ref{tab:mwe-swe} in the case of grammar types like adverbs (ADV), nouns (NOU), noun phrases (NPH), proper names (NPR) or entities (ENT); while others like conjunctions (CON) or propositions (PRO) are domain-independent. 

We quantify the impact of each individual type of linguistic annotation and their possible combinations on the resulting embeddings following different approaches. First, we do this intrinsically through word similarity and relatedness, as well as analogical reasoning tasks. In this regard, we observe that current word similarity and analogy benchmarks, which have become the standard means for the intrinsic evaluation of word embeddings, are of little help in domains like science, which have highly specialized vocabularies. This observation highlights the need for domain-specific benchmarks. Second, we also evaluate extrinsically in two additional tasks: a word prediction task against scientific corpora extracted from SciGraph and Semantic Scholar~\cite{Ammar2018ConstructionOT}, as well as general-purpose corpora like Wikipedia and UMBC~\cite{UMBCEBIQUITYCORESemanticTextualSimilaritySystems}, and a classification task based on the SciGraph taxonomy of scientific disciplines.

\begin{table}[t]
\small
  \caption{\small Multi-word expressions (mwe) vs. single-word expressions (swe) per grammar type (PoS) in SciGraph. Number of swe and mwe occurrences are in millions.}
  \label{tab:mwe-swe}
\begin{center}
\begin{tabular}{lccc}
\hline
\textbf{PoS} & \textbf{example} & \textbf{\#swe} & \textbf{\#mwe} \\ %& \textbf{mwe/swe} \\
\hline
ADJ & single+dose & 104.7 & 0.514 \\ %& 0.005\\
ADV & in+situ & 21 & 1.98 \\ %& 0.094\\
CON & even+though & 34.6 & 3.46 \\ %& 0.1\\
ENT & august+2007 & 1.5 & 1.24 \\ %& 0.82\\
NOU & humid+climate & 216.9 & 15.24 \\ %& 0.07\\
NPH & Ivor+Lewis & 0.917 & 0.389 \\ %& 0.42\\
NPR & dorsal+ganglia & 22.8 & 5.22 \\ %& 0.22\\
PRO & other+than & 13.89 & 0.005 \\ %& 0.0\\
VER & take+place & 69.39 & 0.755 \\ %& 0.01\\
\hline
\end{tabular}
\end{center}
\end{table}

The remainder of this paper is structured as follows: Section~\ref{sec:rel} contains an account of related work. Section~\ref{sec:theo} describes our theoretical framework, which addresses different ways to capture various types of linguistic and semantic information in a single embedding space. In this regard, we focus on Vecsigrafo~\cite{Denaux2019Vecsigrafo}, which allows jointly learning word, lemma and concept embeddings using different tokenization strategies, as well as linguistic annotations. Section~\ref{sec:exp} presents our experiments, where we identify significant findings. For example, embeddings learned from corpora enriched with concept information consistently improve word similarity and analogical reasoning results, while the latter additionally tends to benefit from surface form embeddings. Other results indicate e.g. that using grammatical information to generate scientific embeddings significantly improves the precision of classification but in contrast penalizes the quality of the embeddings for word prediction. Section~\ref{sec:disc} provides a more detailed discussion based on our findings. Finally, Section~\ref{sec:conc} concludes the paper.

\section{Related work}
\label{sec:rel}
The algorithms to learn word embeddings were originally categorized by  Pennington et al.~\cite{pennington2014glove} as global matrix factorization (count-based) approaches \cite{Levy2014EmbMatrixFac,pennington2014glove,shazeer2016swivel} and local context window methods (prediction) \cite{mikolov2013a,bojanowski2016enriching}. Levy et al.~\cite{Levy2014EmbMatrixFac} proved that local context window methods implicitly factorize a word-context matrix. % whose cells are the pointwise mutual information (PMI) of the corresponding word and context pairs, shifted by a global constant. 
Although successfully applied in a variety of problems, word embedding algorithms usually work at a word or n-gram level. Context-free models such as word2vec or GloVe thus generate a single word embedding representation for each word in the vocabulary (“jaguar” would have the same embedding either as an animal or as a car brand). Plus, they ignore multi-word expressions, grammatical and word-sense information, which may be useful e.g. to deal with polysemy and homonymy. 

Recently, considerable effort has shifted towards contextual word embeddings using neural language models based on e.g. bi-directional LSTMs~\cite{PetersELMO2018} or transformer encoders~\cite{devlin2018bert}. Although these are without a doubt a powerful tool in most modern NLP architectures, we are still uncertain as to what information they actually capture. Recent studies, such as~\cite{jawahar-etal-2019-bert, tenney-etal-2019-bert}, conclude that language models like BERT seem to learn surface features at the bottom layers, syntactic features in the middle, and semantic features at the top, following a compositional tree-like structure. In addition, other contributions have shown that BERT and ELMo output embeddings and hidden states encode parse tree information \cite{hewitt-manning-2019-structural}, while BERT can also attend at linguistic notions of syntax and coreference ~\cite{Clark2019WhatDB}. However, in spite of such efforts the linguistic and semantic information thus captured is still hard to interpret logically. In this paper, we use explicit linguistic and semantic features extracted from text and structured knowledge graphs and quantify their impact on the resulting embeddings.

Several approaches focus on learning embeddings that leverage structured graph representations like WordNet~\cite{Miller:1995:WLD:219717.219748} to capture the semantics represented explicitly in such resources. Among such approaches, TransE~\cite{bordes2013transE}, HolE~\cite{nickel2016HolE}, RDF2Vec~\cite{ristoski2016rdf2vec}, Graph Convolutions~\cite{schlichtkrull2018rgcn} and ComE+~\cite{Cavallari2019CommunityKGEmbs} (see~\cite{Ji2020SurveyKGs} for a recent review) directly encode the representations contained in the knowledge graph, which is typically an already condensed and filtered version of real-world data. A related research direction, with approaches like NASARI~\cite{camacho2016nasari}, SW2V~\cite{mancini2017sw2v} and Vecsigrafo~\cite{Denaux2019Vecsigrafo}, proposes to co-train word and concept embeddings based not only on knowledge graphs, but also on text corpora through lexical specificity or word-sense disambiguation. This has several advantages against using just the knowledge graph, including additional coverage of the different domains of interest and the additional signal captured from the linguistic information contained in the text. 

The main contribution of this paper is a comprehensive experimental study on the quality of embeddings generated from a scientific corpus using additional, explicit annotations with linguistic and semantic information extracted from the text. To this purpose, we follow the approach proposed by Vecsigrafo. However, in contrast to the experiments described in~\cite{Denaux2019Vecsigrafo}, which used general domain corpora, here we rely on domain-specific text from scientific publications and include the use of grammatical information (part-of-speech) in our experimentation. Also, our study covers all the possible combinations of the resulting embeddings, including those related to space-separated tokens, surface forms, lemmas, concepts, and grammatical information. Finally, we add an extrinsic evaluation task on scientific text classification to further assess the quality of the embeddings.

\section{Theoretical framework}
\label{sec:theo}
Next, we describe our approach to jointly learn embeddings using linguistic annotations on a text corpus, the strategies to mix and merge such embeddings, and the evaluation tasks used in this work. 

\subsection{Vecsigrafo: corpus-based word-concept embeddings}
\begin{figure*}[htb!]
\centering
\includegraphics[width=0.9\textwidth]{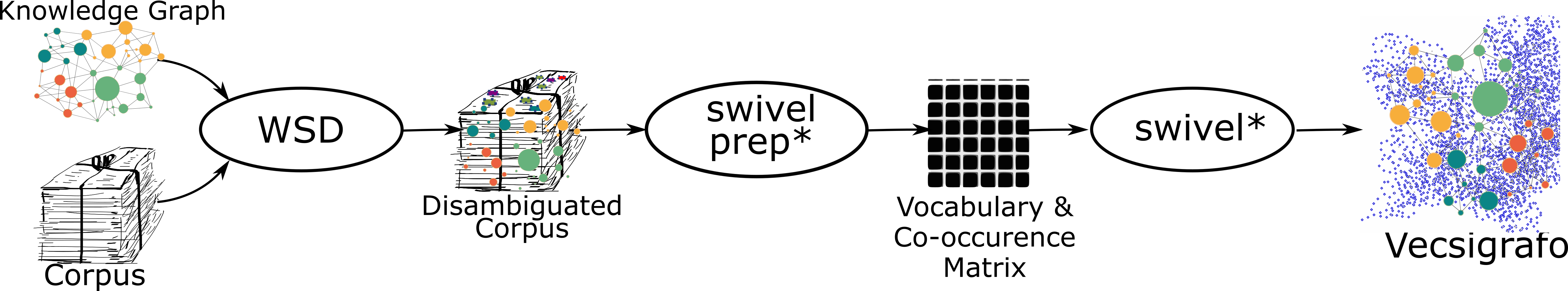}
\caption{Vecsigrafo generation process (Figure taken from \cite{Denaux2019Vecsigrafo}).}
\label{fig:vecsigrafo}
\end{figure*}

Vecsigrafo~\cite{Denaux2019Vecsigrafo} is a method for jointly learning concept and word embeddings from a text corpus. Like word2vec, it uses a sliding window over the text to predict pairs of center and context words. Unlike word2vec, which relies on a simple tokenization strategy based on space-separated tokens (\textit{t}), in Vecsigrafo the aim is to learn embeddings also for linguistic annotations, including surface forms ($\textit{sf} \in SF$), lemmas ($\textit{l} \in L$), grammar information ($\textit{g} \in G$) and concept identifiers ($\textit{c} \in C$) from a knowledge graph. To derive such additional information from a text corpus, in Vecsigrafo the use of a word-sense disambiguation pipeline (see Figure \ref{fig:vecsigrafo}) is required to extract from the original text the corresponding surface forms, lemmas, grammar information, e.g. part-of-speech tags, and concept identifiers. 

\begin{table*}[htb!]
  \centering
  \footnotesize
   \caption{Example of research paper title and linguistic annotations added to jointly learn vecsigrafo embeddings for surface forms, lemmas and concepts. Vecsigrafo learns embeddings for each instance of the annotation types. Each box is a set of annotations with the following pattern: <surface form>|<lemma>|<concept>. Lemmas use the prefix "lem\_" and concepts "en\%".}
    \begin{tabular}{c c}
    \toprule
    %Type & Tokens and linguistic annotations \\
    %\midrule
    
    %
    \texttt{t} & \begin{minipage}[t]{0.85\textwidth}
    \texttt{a novel wasp gene mutation in a chinese boy with wiskott-aldrich syndrome} 
    \end{minipage} \\
    \\
    
    \texttt{sf\_l\_c} & \begin{minipage}[t]{0.85\textwidth}\texttt{a \fbox{novel|lem\_novel|en\%23100008261}  \fbox{wasp|lem\_Wiskott-Aldrich+syndrome+protein} \fbox{gene+mutation|lem\_gene+mutation|en\%23101415380} in a  \fbox{chinese|lem\_Chinese|en\%2398003} \fbox{boy|lem\_boy|en\%2346011} with \fbox{wiskott-aldrich+syndrome|lem\_Wiskott-Aldrich+syndrome}} 
    \end{minipage} \\
    
    \bottomrule 
    \end{tabular}
    \label{tab:annotations}
\end{table*}

In contrast to simple tokenization strategies based on space separation, surface forms are the result of a grammatical analysis where one or more tokens can be grouped using the part of speech information e.g. in noun or verb phrases. Surface forms can include multi-word expressions that refer to concepts, e.g. "Cancer Research Center". A lemma is the base form of a word, e.g. surface forms "is", "are", "were" all have lemma "be". 
Due to the linguistic annotations that are included in addition to the original text, the word2vec approach is no longer directly applicable. The annotated text is a sequence of words and annotations. Thus, sliding a context window from a central word needs to take into consideration  both words and annotations. For this reason, Vecsigrafo extends the Swivel~\cite{shazeer2016swivel} algorithm to first use the sliding window to count co-occurrences between pairs of linguistic annotations in the corpus. These are then used to estimate the point-wise mutual information (PMI) between each pair. The algorithm then tries to predict all of the pairs as accurately as possible.

To illustrate how Vecsigrafo works let us analyse the text, "\textit{a novel wasp gene mutation in a chinese boy with Wiskott-Aldrich syndrome}". Swivel splits the text into 12 space-separated tokens, then builds the co-occurrence matrix with these tokens as features in rows and columns. In contrast, Vecsigrafo first enriches the text with the linguistic annotations that are shown in Table \ref{tab:annotations} and then each annotation (surface forms, lemmas, and concepts) is used as a feature in the swivel co-occurrence matrix. Note that Swivel understands "gene" and "mutation" as distinct tokens, while Vecsigrafo identifies them as a single surface form, a lemma and a concept, and then generates embeddings for each of them. More importantly, while for the ambiguous term "wasp" Swivel generates an embedding, Vecsigrafo realizes that in this context the surface form "wasp" refers to the lemma "Wiskott-Aldrich syndrome protein", and encodes this contextual information in the corresponding embeddings.

In the end, Vecsigrafo produces an embedding space: $\Phi=\{(x,e): x \in SF \cup L \cup G \cup C, e \in \mathbf{R}^{n}\} $. That is, for each entry $x$, it learns an embedding of dimension $n$, such that the dot product of two vectors predicts the PMI of the two entries. For the interested reader, a complete description of Vecsigrafo algorithm is available in ~\cite{Denaux2019Vecsigrafo}.

\subsection{Joint embedding space}\label{sec:merging}
To use Vecsigrafo embeddings, we need to annotate a corpus $D$ with the linguistic annotations used to learn the embeddings. An annotated document in $D$ is a sequence of annotation tuples $D_{j}=((sf,l,g,c)_{i})^{|D_{j}|-1}_{i=0}$. Then we need to generate out of $D$ a vocabulary for each linguistic annotation $V(D)=\{v:v \in SF_{D} \cup L_{D} \cup G_{D} \cup C_{D} \}$, and represent any annotated document as a finite sequence of items in this vocabulary $S(D_{j})= (sf_{i},l_{i},g_{i},c_{i})^{|D_{j}|-1}_{i=0}$. Following this approach, there is a direct mapping between any sequence element in $S(D_{j}))$ and Vecsigrafo embeddings in $\Phi$. %We call this approach single linguistic element vocabulary or \textbf{Single-LEV}. 

In addition, it is possible to merge Vecsigrafo embeddings, such as those corresponding to surface forms and concepts. This is in line with common practice in the literature involving e.g. the combination of conventional and contextualized embeddings~\cite{PetersELMO2018, devlin2018bert} or the creation of  multimodal embedding spaces~\cite{Thoma2017TowardsHC}. Different operations can be applied on the embeddings, including concatenation, average, addition, and dimension reduction techniques such as principal component analysis PCA, singular vector decomposition SVD or neural auto-encoders.

\subsection{Embeddings evaluation}
Embeddings evaluation can be classified in intrinsic and extrinsic methods \cite{schnabel2015evalMethods}. The former assesses whether the embedding space actually encodes the distributional context of words, while the latter makes an assessment according to the performance of a downstream task. The evaluation tasks described in this section are intended to shed light on the quality of the scientific embeddings generated using linguistic annotations and to compare them with regular, space-separated token embeddings. Since the embeddings are learned from a scientific corpus, it is not expected that they outperform the state of the art in intrinsic tasks, such as word similarity and relatedness \cite{LASTRADIAZ2019645} or analogical reasoning, where datasets comprise mostly terms and relations from general lexicon with little overlap with ours.

\textbf{Word similarity and relatedness } \cite{Resnik:1995:UIC:1625855.1625914} is an intrinsic evaluation method that assesses whether the embedding space captures the semantic relatedness between words. To evaluate the embeddings in this task, we compute the correlation between an embedding similarity measure, usually the cosine function, and human similarity or relatedness scores. In the word embeddings literature, Spearman's rho correlation is widely used \cite{pennington2014glove,shazeer2016swivel}.  However, Pearson correlation, alone or in combination with Spearman's, is reported in several papers on word similarity and relatedness \cite{MihalceaSemRelatedness, camacho2015nasari,LASTRADIAZ2019645} as a way to evaluate the strength of such similarity and relatedness between word pairs and the correct ranking. In this paper, we adhere to the latter and report both correlations' metrics.

To work with the concept embeddings produced by Vecsigrafo, we use a concept-based word similarity measure. Essentially, each pair of words is mapped to concepts, using the most frequent word meaning as reference, and similarity is calculated on the corresponding concept embeddings.  %To do so, we calculate the maximum similarity between the concepts associated to the initial pair of words. %To denote the experiments where the concept-based word similarity was used we use the suffix \_c to the embedding label. 

\textbf{Word analogy} \cite{mikolov2013b} is another intrinsic evaluation method where the goal is to find x such that the relation x:y resembles a sample relation a:b by operating on the corresponding word vectors. 

\textbf{Word prediction}, introduced in~\cite{Denaux2019Vecsigrafo}, proposes the use of a test corpus to model how accurately the resulting embeddings can predict a linguistic annotation (\textit{t}, \textit{sf}, \textit{l}, \textit{c} or \textit{g}). This approach essentially emulates the word2vec skipgram loss function on a test corpus unseen during training, iterating through the sequence of tokens in a corpus using a context window. For each center word or linguistic annotation, we calculate the (weighted) average of the embeddings for the context tokens and compare it to the embedding for the center word using cosine similarity. If the cosine similarity is close to 1, the center word is being correctly predicted based on its context. We use this to plot the similarities for the whole embedding vocabulary, providing insights about possible disparities in the quality of common and uncommon words, as well as the applicability of the resulting embeddings to a different corpus.

\textbf{Text classification:} Since word embeddings are usually at the input of neural text classifiers, the performance of the latter can be used as an indicator of the fitness of the former. In contrast to models frequently used for text classification~\cite{Manning:2008:IIR}, such as Naive Bayes and Support Vector Machines (SVM)~\cite{Wang2012LinearClassifiers} which work in a highly dimensional space, deep neural networks work on a compressed space thanks to the use of fixed length embeddings. Neural networks with convolutional layers CNN \cite{lecun1998gradient} using a single layer  \cite{Collobert2008CNNNLP,kim2014convolutional} or multiple-layers \cite{dos2014CNNTextClassification, Zhang2015CCN} have shown state of the art performance for NLP tasks without the need of hand-crafted features \cite{Collobert2008CNNNLP}, one of the main limitation of traditional approaches. %The convolutional layers slide filters (or kernels) across the input data automatically learning features, and these layers can be stacked allowing feature composition with a increased level of abstraction. 

\section{Experimental work}
\label{sec:exp}
This section describes the corpus, the embeddings learned from the linguistic annotations, and the experiments carried out to gain insights on their usefulness in the evaluation tasks. 

\subsection{Learning scientific embeddings with Vecsigrafo}
The scientific corpus used in the following experiments is extracted from SciGraph~\cite{HammondPT17:SciGraphModeling}, a knowledge graph for scientific publications. From it, we extract titles and abstracts of articles and book chapters published between 2001 and 2017. The resulting corpus consists roughly of 3.2 million publications, 1.4 million distinct words, and 700 million tokens. 

Next, we use Expert.ai's Natural Language API to parse and disambiguate the text and add the linguistic annotations associated to each token or group of tokens. To this purpose, the API relies on its own knowledge graph, called Sensigrafo, which encodes linguistic information in a way similar to WordNet and follows a rule-based approach to word-sense disambiguation. Sensigrafo contains approximately 400K lemmas and 300K concepts interlinked via 61 relation types and is available in 14 languages. In addition to Sensigrafo, Vecsigrafo has been successfully trained on several knowledge graphs\footnote{{\em Tutorial on Hybrid Techniques for Knowledge-based NLP}\\ \url{http://hybridnlp.expertsystemlab.com/tutorial}}, including WordNet. For convenience, in this work we use Sensigrafo as the reference knowledge graph in our experiments. Furthermore, it should be noted that any other NLP toolkit would have been equally suitable as long as the generation of the linguistic annotations used in this work are supported. 

\begin{table}[t!]
    \centering
     \caption{Account of token types and embeddings learned from the text in the title and abstracts from research articles and book chapters in English published between 2001 to 2017 and available in SciGraph.}
  \resizebox{\columnwidth}{!}{%
    \begin{tabular}{ccccc}
    \toprule
    \textbf{Algorithm} & \textbf{Types} & \textbf{Total} & \textbf{Distinct} & \textbf{Embeddings} \\
    \midrule
    Swivel & \textit{t}  & 707M & 1,486,848 & 1,486,848 \\
    \multirow{4}{*}{Vecsigrafo} & \textit{sf}  & 805M & 5,090,304 & 692,224 \\
    & \textit{l}  & 508M & 4,798,313 & 770,048 \\
    & \textit{g} & 804M & 25   & 8 \\
    & \textit{c} & 425M & 212,365 & 147,456 \\
    \bottomrule
    \end{tabular}%
  }

  \label{tab:corpus}%
\end{table}%

The corpus parsing and annotations generated by the Natural Language API are reported in Table \ref{tab:corpus}. For space-separated tokens, we learned an initial set of embeddings using Swivel, which is the reference algorithm for Vecsigrafo. For the remaining linguistic annotations (\textit{sf}, \textit{l}, \textit{g}, \textit{c}), as well as their combinations of size 2 and 3, we learned their embeddings using Vecsigrafo. The quantitative difference between the number of distinct annotations and actual embeddings is due to filtering. Following previous findings~\cite{Denaux2019Vecsigrafo}, we filtered out elements with grammar type articles, punctuation marks or auxiliary verbs and generalized tokens with grammar type entity or person proper noun, replacing the original token with special tokens \textit{grammar\#ENT} and \textit{grammar\#NPH}, respectively. 

\subsection{Word similarity and relatedness}

In Table \ref{tab:simdataset}, we report the 16 datasets used in the word similarity evaluation task. Note that we only applied concept similarity in those combinations where \textit{sf} or \textit{l} were not present. Table \ref{tab:simeval} shows the word similarity evaluation results for the Vecsigrafo embeddings in the 16 similarity and relatedness datasets. 

\begin{table}[t!]
\centering
\caption{Similarity and relatedness datasets used for embeddings evaluation.}
   \resizebox{\columnwidth}{!}{
    \begin{tabular}{lllcc}
    \toprule
    \textbf{Dataset} & \multicolumn{1}{p{4.215em}}{\textbf{Main Content}} &
    \multicolumn{1}{p{2.5em}}{\textbf{Rel. Type}} &
    \multicolumn{1}{p{3 em}}{\textbf{Word Pairs}} & \multicolumn{1}{p{3 em}}{\textbf{Raters per Pair}} \\
    \midrule
    The RG-65 \cite{rubenstein1965rg65} & Nouns & Sim. & 65    & 51    \\
    MC-30 \cite{miller1991mc30}         & Nouns & Sim. & 30    & 51  \\
    WS-353-ALL \cite{finkelstein2002ws353} & Nouns & Mix.  & 353 &  13-16  \\
    WS-353-REL  \cite{agirre2009similAndRelatedness} & Nouns & Rel. & 252   & 13-17 \\
    WS-353-SIM  \cite{agirre2009similAndRelatedness} & Nouns & Sim. & 203   & 13-18  \\
    YP-130  \cite{yang2006yp130}  & Verbs & Rel. & 130  & 6     \\
    VERB-143  \cite{baker2014Verb144} & \multicolumn{1}{p{3.57em}}{Conjugated Verbs} & Rel. & 143   & 10     \\
    SIMVERB3500 \cite{gerz2016simverb3500} & \multicolumn{1}{p{3.57em}}{lemmatized Verbs} & Sim. &3500  & 10     \\
    MTurk-287  \cite{radinsky2011mturk287} & Nouns & Rel. & 287   & 23     \\
    MTurk-771 \cite{koehn2005epc} & Nouns & Rel. & 771 & 20 \\
    MEN-TR-3K \cite{bruni2012men3k} & Nouns & Rel. & 3000   & 10  \\
    SIMLEX-999-Adj \cite{hill2014simlex999} & Adjectives & Sim.  & 111   & 50     \\
    SIMLEX-999-Nou \cite{hill2014simlex999} & Nouns & Sim. & 666   & 50     \\
    SIMLEX-999-Ver \cite{hill2014simlex999} & Verbs &Sim. &222   & 50     \\
    RW-STANFORD \cite{luong2013rw} & \multicolumn{1}{p{3.57em}}{Noun, Verbs, Adj} & Rel. &2034  & 10     \\
    SEMEVAL17 \cite{camacho2017semeval17t2} & \multicolumn{1}{p{3.57em}}{Named Entities, MWE} &  Rel. & 500   & 3       \\
    \bottomrule
    \end{tabular}%
    }
  \label{tab:simdataset}%
\end{table}%

\begin{table*}[ht!]
  \centering
  \caption{Spearman's rho for a subset of all the similarity datasets and the final learned vecisgrafo and swivel embeddings. Results are sorted in descending order on the harmonic mean of Spearman's rho and Pearson correlation.}
  \resizebox{\textwidth}{!}{
    % Table generated by Excel2LaTeX from sheet 'formatted table'
    \begin{tabular}{r|rrrrrrrrrrrrrrrr|rrr}
    \toprule
    \begin{sideways}\textbf{Embeddings}\end{sideways} & \begin{sideways}\textbf{MC-30}\end{sideways} & \begin{sideways}\textbf{MEN-TR-3k}\end{sideways} & \begin{sideways}\textbf{MTurk-287}\end{sideways} & \begin{sideways}\textbf{MTurk-771}\end{sideways} & \begin{sideways}\textbf{RG-65}\end{sideways} & \begin{sideways}\textbf{RW-STANFORD}\end{sideways} & \begin{sideways}\textbf{SEMEVAL17}\end{sideways} & \begin{sideways}\textbf{SIMLEX-999-Adj}\end{sideways} & \begin{sideways}\textbf{SIMLEX-999-Nou}\end{sideways} & \begin{sideways}\textbf{SIMLEX-999-Ver}\end{sideways} & \begin{sideways}\textbf{SIMVERB3500}\end{sideways} & \begin{sideways}\textbf{VERB-143}\end{sideways} & \begin{sideways}\textbf{WS-353-ALL}\end{sideways} & \begin{sideways}\textbf{WS-353-REL}\end{sideways} & \begin{sideways}\textbf{WS-353-SIM}\end{sideways} & \begin{sideways}\textbf{YP-130}\end{sideways} & \begin{sideways}\textbf{Avg. Spearman}\end{sideways} & \begin{sideways}\textbf{Avg. Pearson}\end{sideways} & \begin{sideways}\textbf{Harmonic mean}\end{sideways} \\
    \midrule
    l\_c  & 0.642 & \textbf{0.645} & 0.556 & \textbf{0.526} & 0.561 & \textbf{0.359} & 0.553 & 0.421 & 0.330 & 0.078 & 0.162 & \textbf{0.265} & 0.538 & \textbf{0.539} & 0.553 & 0.439 & \textbf{0.448} & 0.443 & \textbf{0.445} \\
    sf\_l\_c & \textbf{0.670} & 0.612 & 0.473 & 0.487 & 0.614 & 0.299 & 0.568 & 0.366 & 0.298 & 0.184 & 0.152 & 0.189 & 0.536 & 0.535 & 0.580 & 0.447 & 0.438 & \textbf{0.446} & 0.442 \\
    sf\_c & 0.630 & 0.635 & 0.478 & 0.486 & 0.573 & 0.328 & 0.549 & 0.417 & 0.289 & 0.117 & 0.148 & 0.248 & \textbf{0.541} & 0.526 & \textbf{0.588} & 0.357 & 0.432 & 0.440 & 0.436 \\
    l\_g\_c & 0.596 & 0.629 & 0.518 & 0.517 & 0.521 & 0.281 & 0.544 & \textbf{0.455} & 0.308 & 0.176 & 0.174 & 0.138 & 0.473 & 0.478 & 0.515 & 0.458 & 0.424 & 0.426 & 0.425 \\
    l     & 0.649 & 0.607 & 0.519 & 0.496 & \textbf{0.697} & 0.292 & 0.520 & 0.307 & 0.298 & 0.096 & 0.129 & 0.186 & 0.486 & 0.455 & 0.553 & 0.332 & 0.414 & 0.411 & 0.413 \\
    c     & 0.653 & 0.612 & \textbf{0.558} & 0.449 & 0.565 & 0.299 & \textbf{0.583} & 0.316 & 0.344 & 0.119 & 0.248 & -0.243 & 0.471 & 0.442 & 0.571 & 0.507 & 0.406 & 0.402 & 0.404 \\
    sf\_l & 0.539 & 0.585 & 0.483 & 0.423 & 0.545 & 0.251 & 0.501 & 0.382 & 0.258 & 0.159 & 0.138 & 0.169 & 0.510 & 0.509 & 0.519 & 0.363 & 0.396 & 0.400 & 0.398 \\
    sf\_g\_c & 0.532 & 0.620 & 0.521 & 0.456 & 0.469 & 0.280 & 0.542 & 0.406 & 0.305 & 0.141 & 0.146 & 0.156 & 0.452 & 0.447 & 0.476 & 0.266 & 0.388 & 0.399 & 0.394 \\
    sf    & 0.515 & 0.582 & 0.475 & 0.439 & 0.440 & 0.282 & 0.529 & 0.384 & 0.303 & 0.130 & 0.153 & 0.121 & 0.472 & 0.472 & 0.490 & 0.363 & 0.384 & 0.380 & 0.382 \\
    g\_c  & 0.446 & 0.584 & 0.490 & 0.405 & 0.480 & 0.275 & 0.571 & 0.323 & \textbf{0.350} & \textbf{0.201} & \textbf{0.267} & -0.228 & 0.420 & 0.409 & 0.509 & \textbf{0.508} & 0.376 & 0.384 & 0.380 \\
    sf\_l\_g & 0.510 & 0.564 & 0.494 & 0.401 & 0.327 & 0.237 & 0.446 & 0.364 & 0.249 & 0.144 & 0.110 & 0.124 & 0.429 & 0.462 & 0.462 & 0.318 & 0.353 & 0.360 & 0.356 \\
    l\_g  & 0.536 & 0.552 & 0.535 & 0.430 & 0.523 & 0.219 & 0.454 & 0.228 & 0.251 & 0.095 & 0.119 & 0.001 & 0.364 & 0.420 & 0.385 & 0.385 & 0.344 & 0.346 & 0.345 \\
    sf\_g & 0.502 & 0.524 & 0.460 & 0.395 & 0.345 & 0.197 & 0.436 & 0.239 & 0.231 & 0.091 & 0.102 & 0.109 & 0.400 & 0.421 & 0.464 & 0.332 & 0.328 & 0.329 & 0.328 \\
    t     & 0.200 & 0.332 & 0.323 & 0.220 & 0.483 & 0.218 & 0.051 & 0.337 & 0.164 & 0.180 & 0.172 & 0.107 & 0.155 & 0.101 & 0.187 & 0.308 & 0.221 & 0.279 & 0.247 \\
    \bottomrule
    \end{tabular}%
   }
    \label{tab:simeval}
\end{table*}%

From the data, we observe clear evidence that the embeddings derived from linguistic annotations, individually or jointly learned, perform better in the similarity task than using token embeddings. An individual analysis of each linguistic annotation shows that \textit{l} embeddings outperform \textit{c} embeddings, which in turn are superior to \textit{sf} embeddings. Most of the similarity datasets in Table \ref{tab:simdataset} contain mainly nouns and lemmatized verbs. Semantic and lemma embeddings seem to be better representations for these datasets, since they condense different words and morphological variations in one representation. Surface forms, on the other hand, may generate several embeddings for the different morphological variations of words, which might not be necessary for this task and can hamper the similarity estimation. 

Regarding Vecsigrafo embeddings jointly learned for two or more linguistic annotations, semantic embeddings \textit{c} improve the results in all the combinations where they were used. Concept embeddings enhance the lexical representations based on surface forms and the grouped morphological variations represented by lemmas in the similarity task. On the other hand, grammatical information \textit{g} seems to hamper the performance of Vecsigrafo embeddings except when they are jointly learned with the semantic embeddings \textit{c}. If we focus the analysis on the top 5 results in Table \ref{tab:simeval}, \textit{l} and \textit{c} embeddings, jointly or individually learned, are the common elements in the embedding combinations that perform best in the similarity task. On the other hand, the performance of the embeddings from surface forms \textit{sf} and related combinations is at the bottom of the table. In general, the embeddings based on \textit{l} and \textit{c} seem to be better correlated to the human notion of similarity than embeddings based on \textit{sf} and \textit{g}.

To summarize, the results obtained in the similarity task show evidence that the embeddings  benefit from the additional semantic information contributed by the linguistic annotations. Lexical variations presented at the level of surface forms are better representations in this regard than space-separated tokens. The conflation of these variations in a base form at the lemma level and their link to an explicit concept increases performance in the prediction of word relatedness. Jointly learned embeddings for surface forms, lemmas, and concepts achieve the best overall similarity results. 

The analysis in~\cite{Denaux2019Vecsigrafo} evaluates, using the same experimental settings, Vecsigrafo embeddings for lemmas and concepts learned from a 2018 dump of Wikipedia, achieving a Spearman's rho of 0.566. This contrasts with the 0.448 achieved by the same embeddings learned from the SciGraph corpus. These results may not be directly comparable due to the difference in size between the two corpora (3B tokens in Wikipedia vs. 700M in SciGraph). Another factor is that Wikipedia is a general-purpose resource, while SciGraph is specialized in the scientific domain. 

\subsection{Analogical reasoning}
For the analogy task, we use the Google analogy test set\footnote{Analogy dataset available at:  \url{https://aclweb.org/aclwiki/Google\_analogy\_test\_set\_(State\_of\_the\_art)}}, which contains 19,544 question pairs (8,869 semantic questions and 10,675 syntactic questions) and 14 types of relations (9 syntactic and 5 semantic). The accuracy of the embeddings in the analogy task is reported in Table \ref{tab:analogy}.  

\begin{table}[h!]
\centering
\caption{Results of the analogy tasks sorted in descending order on the overall results.}
  \resizebox{0.7\columnwidth}{!}{%
    \begin{tabular}{cccc}
    \toprule
          & \multicolumn{3}{c}{\textbf{Analogy Task}} \\
\cmidrule{2-4}    \textbf{Embeddings} & \textbf{Semantic} & \textbf{Syntactic} & \textbf{Overall} \\
    \midrule
    sf\_c & 11.72\% & \textbf{14.17\%} & \textbf{12.94\%} \\
    sf\_g\_c & 14.13\% & 11.65\% & 12.89\% \\
    sf\_l\_c & 9.27\% & 12.10\% & 10.69\% \\
    l\_g\_c & \textbf{14.83\%} & 3.09\% & 8.96\% \\
    sf\_l & 7.18\% & 8.63\% & 7.91\% \\
    l\_c  & 10.47\% & 4.06\% & 7.26\% \\
    sf    & 6.69\% & 6.43\% & 6.56\% \\
    sf\_l\_g & 6.60\% & 5.74\% & 6.17\% \\
    sf\_g & 6.07\% & 3.79\% & 4.93\% \\
    l     & 4.07\% & 1.27\% & 2.67\% \\
    l\_g  & 2.33\% & 0.79\% & 1.56\% \\
    t     & 0.64\% & 0.09\% & 0.37\% \\
    \bottomrule
    \end{tabular}%
     }
  \label{tab:analogy}%
\end{table}%

Along the same lines as the word similarity and relatedness results, in this task the linguistic annotations also generate better embeddings than space-separated tokens. However, surface forms are more relevant than lemmas in this case. Our results indicate that surface form embeddings achieve a positive result in both semantic and syntactic analogies while the performance of lemma embeddings is high for semantic analogies and really low in the case of syntactic analogies. In the latter case, the embeddings generated taking into account the morphological variations of words produce better results than those derived from the base forms. Indeed, some syntactic analogies actually require taking into account the information contained in surface forms. For example, capturing analogies like "bright/brighter" or "cold/colder" needs embedding representations for each individual term, whereas in a lemma-based approach only the terms "cold" and "bright" would be represented.

%For example, some syntactic analogies actually require to work at the surface form level e.g., "bright - brighter cold-colder" since the lemmas of \textit{brighter} and \textit{colder} are \textit{bright} and \textit{cold} respectively and therefore in the lemma embedding space there is no representation for \textit{brighter} and \textit{colder}.

The highest accuracy is achieved by jointly learning \textit{sf} and \textit{c} embeddings. In  fact, these two linguistic annotations are among the top 3 results along with \textit{g} or \textit{l}. As in word similarity, semantic embeddings \textit{c} improve every combination in which they are used. If we focus on the semantic analogies, we can see that \textit{l}, \textit{g} and \textit{c} reach the highest accuracy. Nevertheless, the performance of this combination is very poor on syntactic analogies given that \textit{l} do not include morphological variations of words that are heavily represented in the syntactic analogies. In general, the worst results are obtained when the combinations do not include \textit{c}.  

In Shazeer et al.~\cite{shazeer2016swivel}, Swivel embeddings learned from Wikipedia and Gigaword achieved an analogy accuracy in the same dataset used in these experiments of 0.739, while the best result reported in our analysis is 0.129. As in the similarity and relatedness evaluation, the smaller size and domain specificity of the SciGraph corpus seem to have a negative impact on the results obtained in these generic evaluation tasks. 

\subsection{Word prediction}
We select 17,500 papers from Semantic Scholar, which are not in our SciGraph corpus, as the unseen text on which we try to predict the embeddings for certain words (or linguistic annotations) according to the context. We applied the same annotation pipeline to the test corpus so that we can use the embeddings learned from SciGraph. As baselines we used pre-trained embeddings learned for linguistic annotations that we derived from \emph{generic}, i.e. non-scientific, corpora: a dump of Wikipedia from January 2018 containing 2.89B tokens and UMBC~\cite{han2013umbc}, a web-based corpus of 2.95B tokens.

\begin{figure}[t]
  \centering
  \includegraphics[width=0.8\linewidth]{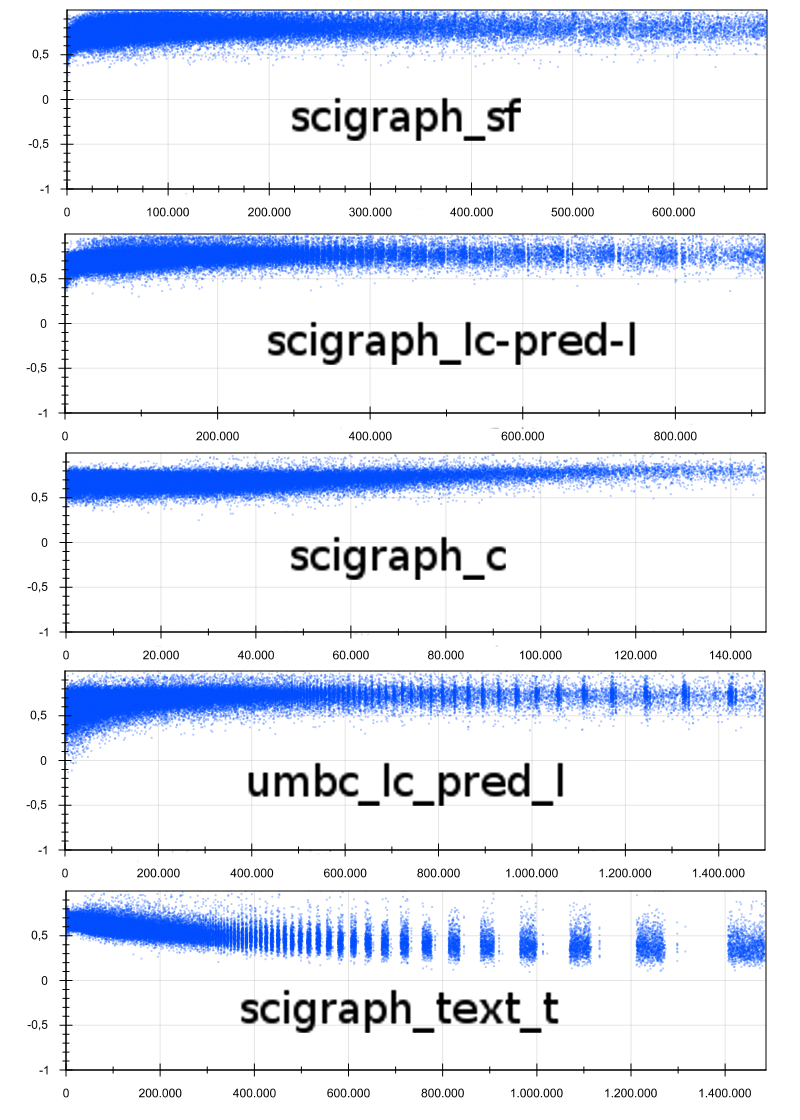}
  \caption{Embedding prediction plots. The horizontal axis aligns with the terms in the vocabulary sorted by frequency. The vertical axis is the average cosine similarity (which can range from 0 to 1) between the predicted embedding on the Semantic Scholar corpus and the actual learned embedding from SciGraph. According to this task the closest the curve draw by the plotted dots to 1 the better. Blank spaces indicates lack of coverage}
  \label{fig:wordpred}
\end{figure}

The overall results are shown in Table~\ref{tab:wordpred}. We also plot the full results for some of the prediction tasks in Figure~\ref{fig:wordpred}. The results show a clear pattern: embeddings learned from linguistic annotations significantly outperform the plain space-separated token embeddings (\textit{t}). That is, the cosine similarity between the predicted embedding for the linguistic annotation on Semantic Scholar and the actual embeddings learned from SciGraph is higher. Recall that the predicted embedding is calculated by averaging the  embeddings of the words or their annotations in the context window.

In general, embeddings learned from linguistic annotations on SciGraph are better at predicting embeddings in the Semantic Scholar corpus. In this case, it was easier to predict surface form embeddings, followed by lemma and concept embeddings. We assume this is because \textit{sf} annotations also contain morphological information. Similarly, concept information may be more difficult to predict due to possible disambiguation errors or because the specific lemma used to refer to a concept may still provide useful information for the prediction task.

\begin{table}[tb!]
\centering
 \caption{Prediction results of embeddings for linguistic annotations.}
  \resizebox{\columnwidth}{!}{%
\begin{tabular}{lllrrrr}
\toprule
\textbf{Corpus} & \textbf{Embed.} & \textbf{pred} & \multicolumn{1}{p{3.215em}}{\textbf{Cosine Sim.}} & \textbf{\#tokens} & \multicolumn{1}{p{3.215em}}{\textbf{Out-of-voc oov}} & \textbf{oov \%}\\
%     &        &               & \textbf{sim.} &                & \textbf{(oov)}        &   \\
\midrule
SciGraph & sf       & sf & 0.765 & 3,841,378 & 617,094 & 16.064\\
SciGraph & sf\_c     & sf & 0.751 & 3,841,378 & 617,144 & 16.066\\
SciGraph & sf\_ l     & sf & 0.741 & 3,841,378 & 617,121 & 16.065\\
SciGraph & l\_ c      & l  & 0.733 & 2,418,362 & 81,087  &  3.353\\
SciGraph & l        & l  & 0.730 & 2,418,362 & 81,041  &  3.351\\
SciGraph & l\_ c      & c  & 0.690 & 2,012,471 & 1,481   &  0.074\\
SciGraph & c        & c  & 0.690 & 2,012,471 & 1,572   &  0.078\\
Wiki & l\_c & l  & 0.676 & 2,418,362 & 116,056 &  4.799\\
UMBC & l\_c & l  & 0.674 & 2,418,362 & 93,169  &  3.853\\
UMBC & l\_c & c  & 0.650 & 2,012,471 & 1,102   &  0.055\\
Wiki & l\_c & c  & 0.650 & 2,012,471 & 2,216   &  0.110\\
%SciGraph & c\_g & c  & 0.642 & 2,012,471 & 1,572   &  0.078\\
SciGraph & t   & t  & 0.576 & 3,396,730 & 205,154 &  6.040\\
\bottomrule
\end{tabular}
  }
  \label{tab:wordpred}%
\end{table}%

Jointly learning embeddings for \textit{sf} and other annotations (\textit{c} or \textit{l}) produces \textit{sf} embeddings which are slightly harder to predict than when trained on their own. However, jointly learning \textit{l} and \textit{c} embeddings produces better results; i.e., jointly learning lemmas and concepts has a synergistic effect.

When comparing to the baselines, we see that SciGraph-based embeddings outperform both the Wikipedia and UMBC-based embeddings. The out-of-vocabulary (oov) numbers provide an explanation: both baselines produce embeddings which miss many terms in the test corpus, but which are included in the SciGraph embeddings. Wikipedia misses 116K lemmas and UMBC misses 93K, but SciGraph only misses 81K of the lemmas in the test corpus. Wikipedia misses the highest number of concepts (2.2K), followed by SciGraph (1.5K) and UMBC (1.1K); however, despite missing more concepts than UMBC, SciGraph-based embeddings outperform both baselines. 

Manual inspection of the missing words shows that UMBC is missing nearly 14K lemmas (mostly for scientific terms) which the SciGraph-based embeddings contain, such as "negative bias temperature instability", "supragranular", "QHD". Inversely, the SciGraph vocabulary is missing nearly 7K lemmas for non-scientific entities, such as "Jim Collins" and "Anderson School of Management", but also for misspelled words. However, most of the missing words (around 42K) are not found in either UMBC or SciGraph and include very specific metrics such as "40 kilopascal", acronyms ("DB620") and named entities ("Victoria Institute of Forensic Medicine"). We observe a similar pattern when comparing missing concepts.

\subsection{Classification of scientific documents}
Leveraging SciGraph metadata, we formulate a multi-label classification task that aims at predicting one or more of 22 first level categories such as Mathematical Sciences, Engineering or Medical and Health Sciences. To train the classifier, we use a vanilla CNN implementation available in Keras. Such neural network consists of 3 convolutional layers with 128 filters and a 5-element window size, each followed by a max pooling layer, a fully-connected 128-unit ReLU layer and a sigmoid output. Our goal is not to design the best classifier, but to evaluate the different types of embeddings through its performance. 

To evaluate the classifiers we select articles published in 2011 and use ten-fold cross-validation. We define a vocabulary size of 20K elements, and max sequence size of 1000. As baseline, we train a classifier that learns from embeddings generated randomly, following a normal distribution. As upper bound we learn a classifier that is able to optimize the embeddings during task-specific training. In addition, we train classifiers using embeddings generated for \textit{t}, \textit{sf}, \textit{l} and \textit{c}. 

\begin{table}[htbp]
\centering
\caption{Learning results using token-based embeddings generated randomly (baseline) and learned by the convolutional neural network (upper bound), and Swivel embeddings for \textit{t}, \textit{sf}, \textit{l} and \textit{c}.}
\resizebox{0.9\columnwidth}{!}{%
\begin{tabular}{ccccc}
\toprule
%\multicolumn{1}{p{6.145e8m}}{\textbf{Embeddings Generation}} & \textbf{Types} & \textbf{P} & \textbf{R} & \textbf{F-measure} \\ 
\textbf{Embed. Algorithm} & \textbf{Types} & \textbf{P} & \textbf{R} & \textbf{F-measure} \\ 
\midrule
Random Normal                & t    & 0,7596     & 0,6775     & 0,7015$\vartriangle$             \\ 
Learned by CNN               & t    & 0,8062     & 0,767      & 0,7806$\triangledown$           \\ 
\midrule
Swivel         & t                   & 0,8008     & 0,7491     & 0,7679             \\ 
Vecsigrafo         & sf                  & 0,8030     & 0,7477     & 0,7684             \\ 
Vecsigrafo         & \textbf{l}          & 0,8035     & 0,7539     & 0,7728             \\ 
Vecsigrafo         & c                   & 0,7945     & 0,7391     & 0,7583             \\ 
\bottomrule
\end{tabular}%7
}
\label{tab:baseline}
\end{table}

The performance of the classifiers measured in terms of precision, recall and f-measure is shown in Table \ref{tab:baseline}. \textit{l} embeddings contribute to learn the best classifier, followed by \textit{sf} and \textit{t} embeddings. If we focus on f-measure we can see again that grammatical analysis (\textit{l, sf}) performs better than the lexical level (\textit{t}). However, the performance of \textit{c} embeddings for this task is rather low. The low number of \textit{c} embeddings (see Table \ref{tab:corpus}) seems to affect negatively the learning process. On the other hand, all the classifiers, including the one trained with \textit{c},  improve the baseline of random embeddings. However, none of the classifiers reached the performance of the upper bound. 

Next, we train classifiers using Vecsigrafo embeddings from all the 10 possible combinations of the linguistic annotations (\textit{sf, l, g, c}) with size 2 and 3. The classifiers are trained using the merging techniques described in Section \ref{sec:merging}. We combine single embeddings per each linguistic annotation type and merge them using vector operators like average, concatenate, and PCA. %In total we trained 60 classifiers. %In addition, to balance vocabulary richness in Single-LEV, which uses a unique vocabulary in contrast to the other approaches, with either more vocabularies or richer vocabulary entries, we increase the vocabulary size to match the Multiple-LEV overall vocabulary size. Hence we trained an additional 10 classifiers. %We increase the vocabulary of all the combinations, except 2-size combinations where grammar was included (sf\_g, l\_g, g\_c) since the number of grammar embeddings is very low, and learn classifiers using the single-LEV approach. 
The average performance of the classifiers is presented in Table \ref{tab:classifiers}. Each column is sorted according to the reported evaluation metric.

\begin{table}[htb]
  \centering
  \caption{Average performance of the 70 classifiers trained using different strategies to mix and merge Vecsigrafo embeddings. Each column is sorted in descending order of the corresponding metric.}
  \resizebox{0.9\columnwidth}{!}{
    \begin{tabular}{lclclc}
    \toprule
    \multicolumn{2}{c}{\textbf{Precision}} & \multicolumn{2}{c}{\textbf{Recall}} & \multicolumn{2}{c}{\textbf{F-Measure}} \\
    \midrule
    \textbf{Types} & \textbf{Average} $\downarrow$ & \textbf{Types} & \textbf{Average} $\downarrow$& \textbf{Types} & \textbf{Average} $\downarrow$\\
    \midrule
    sf\_l\_g & 0,8126 & sf\_l & 0,7605 & sf\_l & 0,7776 \\
    sf\_g & 0,8091 & l\_c  & 0,7556 & sf\_l\_g & 0,7750 \\
    l\_g  & 0,8090 & sf\_l\_c & 0,7538 & l\_g\_c & 0,7730 \\
    sf\_g\_c & 0,8084 & l\_g\_c & 0,7524 & l\_c  & 0,7728 \\
    l\_g\_c & 0,8073 & sf\_l\_g & 0,7524 & sf\_l\_c & 0,7720 \\
    g\_c  & 0,8060 & sf\_c & 0,7520 & sf\_g\_c & 0,7716 \\
    sf\_l & 0,8056 & sf\_g\_c & 0,7504 & l\_g  & 0,7702 \\
    sf\_l\_c & 0,8031 & l\_g  & 0,7477 & sf\_c & 0,7699 \\
    l\_c  & 0,8021 & sf\_g & 0,7374 & sf\_g & 0,7638 \\
    sf\_c & 0,8006 & g\_c  & 0,7338 & g\_c  & 0,7599 \\
    \bottomrule
    \end{tabular}%
    }
  \label{tab:classifiers}%
\end{table}%

Regarding precision, the top 3 classifiers are learned from different combinations of \textit{sf}, \textit{l} and \textit{g}, indicating that, precision-wise, grammatical information is more relevant than semantic information (\textit{c}). However, note that the common linguistic element in the 6 first classifiers is \textit{g}, even when combining it with \textit{c}. In general, removing \textit{g} embeddings produced the least precise classifiers. This means that grammar information is what makes the difference, regardless of the other linguistic elements. That said, the influence of \textit{g} is enhanced when used in combination with \textit{sf} and \textit{l}. Note that the precision of the top 5 classifiers is better than the upper bound classifier (Table \ref{tab:baseline}), where the embeddings were learned at classification training time, although the linguistic-based embeddings were not learned for this specific purpose. 

The recall analysis shows a different picture since grammar embeddings \textit{g} do not seem to have a decisive role on the performance of the classifier recall-wise, while \textit{c} gain more relevance. \textit{sf} and \textit{l} help in learning the best classifier. The combination of \textit{c} and \textit{l} seems to benefit recall, as seen in 3 of the top 4 classifiers. In contrast  when concepts are combined with \textit{sf} the recall is lower. The fact that \textit{l} embeddings are more directly related to \textit{c} than \textit{sf} seems to make a difference in the recall analysis when these three elements are involved. In general, \textit{g}-based embedding combinations generate classifiers with lower recall.

Finally, the f-measure data shows more heterogeneous results since by definition it is the harmonic mean of precision and recall, and hence the combinations that generate the best f-measure also have a high precision and a high recall. The combination of \textit{sf} and \textit{l} is at the top (best recall), followed by their combination with \textit{g} (best precision). \textit{c} appears in positions 3 to 6 in the ranking; however, when combined solely with \textit{sf} or \textit{g}, the f-measure results are the worst. When combined with at least two other elements, \textit{g} ranks high in f-measure, whereas when \textit{g} is combined with a single linguistic annotation type, the resulting performance of the classifiers is the worst. 

\section{Discussion}
\label{sec:disc}
Although the different evaluation tasks assess the embeddings from different perspectives, in all of them we have observed that the embeddings learned from linguistic annotations outperform space-separated token embeddings. In the intrinsic evaluation tasks, the embeddings learned from linguistic annotations have a different performance, mainly due to the vocabulary used in the evaluation datasets. For example, in the analogical reasoning datasets, syntactic analogies are more represented compared to semantic analogies. Also, most of the syntactic analogies often include morphological variations of the words, which are better covered by surface form embeddings than lemma embeddings, where morphological variations are conflated in a single base form. For example, comparative and superlative analogies like \textit{cheap-cheaper} and \textit{cold-colder} may require morphological variations of the adjective by adding the \textit{-er} and \textit{-(e)st} suffixes. Furthermore, although some adjectives and adverbs may produce irregular forms, the only ones included in the dataset are \textit{good-better}, \textit{bad-worse}, \textit{good-best} and \textit{bad-worst}. 

In contrast, in the similarity and relatedness datasets (see Table \ref{tab:simdataset}) most of the word pairs available for evaluation are either nouns, named entities or lemmatized verbs --only the Verb-143 dataset contains conjugated verbs--. Therefore, most of the word pairs are non-inflected word forms. As a result, concept and lemma embeddings are the most suitable representations for this task, which is in line with the evaluation results reported in Table~\ref{tab:simeval}.  

Remarkably, in the analogy task, surface form embeddings jointly learned with concept embeddings achieve the highest performance. In the similarity and relatedness task, jointly learning concepts and lemma embeddings also improves the lemma embeddings performance in the task. Therefore, including concept embeddings in the learning process of other embeddings generally helps to learn better distributional representations. 

As additional insight gained from the word prediction task, we observe that the embeddings learned from linguistic features seem to better capture the distributional context, as evaluated against unseen text. Surface form and lemma embeddings were the easiest to predict.

Finally, in the text classification task, single embeddings from lemma and surface form annotations were more useful than space-separated token embeddings. Nevertheless, in this task, concept embeddings perform worse mainly due to the low coverage offered by such embeddings with respect to the whole vocabulary. This indicates that the knowledge graph used to annotate the text possibly provides limited coverage of the domain. We also show that jointly learning lemma and surface form embeddings helps to train the best classifier in terms of f-measure and recall. Furthermore, adding grammar embeddings produced the best overall precision.

\section{Conclusions and future work}
\label{sec:conc}
In this paper, we present a novel experimental study on the use of linguistic annotations for learning embeddings from a scientific corpus. Our experiments show in all cases that embeddings learned from linguistic annotations perform better than conventional space-separated tokens in all evaluation tasks. Moreover, we demonstrate that the vocabulary, and its grammatical characteristics, used in each evaluation task is directly related to the quality of the linguistic annotated embeddings.

Based on the results of our study, we can conclude that there is not a single type of linguistic annotation that always produces the best results consistently. However, we identify some patterns that may help practitioners to select the most suitable combinations for each specific task and dataset.

In the similarity task, learning embeddings based on lemmas generally works better given the high rate of nouns, named entities and multi-word expressions contained in those datasets. On the contrary, in the analogy dataset there is a majority of syntactic analogies containing morphological variations of words that were better represented by single surface form embeddings. Jointly learning embeddings for concepts and either lemmas or surface forms improve the quality of each individual type of embeddings across all the evaluation tasks. 

Regarding the extrinsic evaluations, in the word prediction task the embeddings learned from linguistic annotations provide evidence to better capture the distributional context of words in an unseen text than space-separated embeddings. In the classification tasks, jointly learned lemma and surface form embeddings help to train the best classifier and, if grammar embeddings are also used, then the highest precision is achieved. 

As future work, we want to understand the type of linguistic information that is being encoded in the Vecsigrafo embeddings beyond the intrinsic evaluation tasks that we used in this paper. Recently, a set of probes has been released to identify whether the models encode different types of linguistic information \cite{hewitt-manning-2019-structural,tenney-etal-2019-bert}, including parse trees, constituent labelling and co-reference to name a few, that can be applied to Vecsigrafo. Another research path we want to explore is to use linguistic annotations to enhance neural language models like BERT and assess whether such linguistic annotations can also help to learn better language models, improving performance when fine-tuned for various downstream tasks. The idea is to extend the current transformer-based architectures used to learn language models, so that not only word pieces, but also other types of linguistic annotations, such as those discussed in this paper, can be ingested and their effect propagated across the model.

\section*{Acknowledgements}
We gratefully acknowledge the EU Horizon 2020 research and innovation programme under grant agreement No. 825627 (ELG). 
%\section*{References}

\bibliography{biblio}

\end{document}